\documentclass{llncs}
\pdfoutput=1
\usepackage{times}
\usepackage{latexsym}
\usepackage{amsmath}
\usepackage{amssymb}
\usepackage{graphicx}
\usepackage{graphics}
\usepackage{booktabs}
\usepackage{array}
\usepackage{adjustbox}
\usepackage{multirow}
\usepackage{cite}
\usepackage[english]{babel}
\usepackage{caption}
\usepackage{url}
\usepackage{hyperref}
\begin{document}
\title{Pretrained language model transfer on neural named entity recognition in Indonesian conversational texts}
\titlerunning{Pretrained LM transfer on neural Indonesian conversational NER}
%


\author{Rezka Leonandya, Fariz Ikhwantri}
\authorrunning{R. Leonandya and F. Ikhwantri}
%
\institute{Kata Research Team, kata.ai, Jakarta, Indonesia \\
\email{\{rezka,fariz\}@kata.ai}}


%
\maketitle              
\begin{abstract}
Named entity recognition (NER) is an important task in NLP, which is all the more challenging in conversational domain with their noisy facets. Moreover, conversational texts are often available in limited amount, making supervised tasks infeasible. To learn from small data, strong inductive biases are required. Previous work relied on hand-crafted features to encode these biases until transfer learning emerges. Here, we explore a transfer learning method, namely language model pretraining, on NER task in Indonesian conversational texts. We utilize large unlabeled data (generic domain) to be transferred to conversational texts, enabling supervised training on limited in-domain data. We report two transfer learning variants, namely supervised model fine-tuning and unsupervised pretrained LM fine-tuning. Our experiments show that both variants outperform baseline neural models when trained on small data (100 sentences), yielding an absolute improvement of 32 points of test F1 score. Furthermore, we find that the pretrained LM encodes part-of-speech information which is a strong predictor for NER.

\end{abstract}
\section{Introduction} \label{sec:intro}

Named entity recognition (NER), the task of assigning a class to a word or phrase of proper names in text, is an essential ability for conversational agents to have. For example, in food delivery application, an agent needs to acquire information about the customer's food detail and address. NER is all the more challenging on conversational texts because of their noisy characteristics, such as typos, informal word variations, and inconsistent naming in named entities. 
Furthermore, conversational texts are often available in diverse domains and limited amount, making supervised training arduous due to data limitation. To learn from limited data, strong inductive biases are necessary. In this work, we explore transfer learning techniques as a way to help neural models learn and generalize from limited data on NER task in Indonesian conversational texts.

Transfer learning, or sometimes known as domain adaptation, is an important approach in NLP application, especially if one does not have enough data in the target domain. In such scenarios, the goal is to transfer knowledge from source domain with large data to target domain so as to improve the model performance on the target domain and prevent overfitting. Early research in transfer learning, especially with entity recognition in mind, were tackled by feature augmentation \cite{Daum2007FrustratinglyED}, bootstrapping \cite{Wu2009DomainAB}, and even rule-based approach \cite{Chiticariu2010DomainAO}.

Recently, neural networks emerge as one of the most potent tools in almost all NLP applications. Although neural models have achieved impressive advancement, they still require an abundant amount of data to reach good performance. With limited data, neural models generalization ability is severely curtailed, especially across different domains where the train and test data distributions are different \cite{Lake2017StillNS}. Therefore, transfer learning becomes more critical for neural models to enable them to learn in data-scarce settings.

Transfer learning in NLP is typically done with two techniques, namely parameter initialization (INIT) and multi-task learning (MULT). INIT approach first trains the network on source domain and directly uses the trained parameters to initialize the network on target domain \cite{Lee2018TransferLF}, whereas MULT simultaneously trains the network with samples from both domains \cite{Aguilar2017AMA}. Recently, INIT approaches were made highlight by the incorporation of pretrained language models \cite{Peters2018DeepCW, Ruder2018UniversalLM, Radford2018ImprovingLU} to neural models, reaching state-of-the-art performance across various NLP tasks.

Indonesian NER itself has attracted many years of research, from as early as using a rule-based approach \cite{Budi2005NIL} to more recent machine learning techniques, such as conditional random field (CRF) \cite{Luthfi2014BuildingAI, Leonandya2015ASA, Taufik2016NamedER}, and support vector machine (SVM) \cite{Suwarningsih2014ImNERIM, Aryoyudanta2016SemisupervisedLA}. The latest research was done by \cite{Kurniawan2018EmpiricalEO} where they investigated neural models performance with word-level and character-level features in Indonesian conversational texts.

In this paper, we apply and evaluate a recent technique of transfer learning, namely language model pretraining. We use the pretrained LM to extract additional word embedding input representations for the neural sequence labeler in Indonesian conversational texts. The work in this paper is organized as follows: We first train a language model on generic domain unlabeled Indonesian texts $\mathbb{U}$ (e.g., Wikipedia). We then use a smaller domain-specific source corpus $\mathbb{S}$ (e.g., task-oriented conversational texts) to either: (a) fine-tune the pretrained LM or (b) train a neural sequence labeler using the pretrained LM's representation as additional input. If we proceed with (a), then the next step is to train a neural sequence labeler on the target domain $\mathbb{T}$ (e.g., small-talk conversational texts) using fine-tuned LMs representation as additional input. If we proceed with (b), then the next step is to fine-tune the neural sequence labeler on the target domain corpus $\mathbb{T}$. We evaluate and compare our approach with other models, namely a neural sequence labeler without LM pretraining and a multi-task approach trained on varying amount of training data from $\mathbb{T}$.


\section{Methodology} \label{sec:metho}
To allow models to learn from small conversational data, we introduce two variants of three-step training procedure. Both variants use additional input of word embedding representations derived from a bidirectional LSTM language model (biLM). The word embedding representations used in this paper is ELMo \cite{Peters2018DeepCW}.

\subsection{Deep Bidirectional Language Models}
ELMo derived from pretrained biLM \cite{Peters2018DeepCW} works exceptionally well in practice, obtaining state-of-the-art performance in six challenging natural language understanding tasks. The biLM is trained by jointly maximizing the log likelihood of the forward and backward language models:
\begin{gather*}
    \sum_{k=1}^{N} (\log p(t_k | t_1,...,t_{k-1};\overrightarrow{\Theta})) + 
    (\log p(t_k | t_{k+1},...,t_{N};\overleftarrow{\Theta} ))
\end{gather*}
where $N$ is the sequence length, $t_k$ is the token at timestep $k$, and $\overrightarrow{\Theta}$ and $\overleftarrow{\Theta}$ are the forward and backward LSTM parameters, respectively.

To compute ELMo, first a set of representations is collected from the biLM. For each token $t_k$, we collect $2L + 1$ representations from the biLM:
\begin{align*}
    R_k &= \{x_k, \overrightarrow{h}_{k,j}, \overleftarrow{h}_{k,j} | j = 1,..,L\}
\end{align*}
where $x_k$ is the context-independent token representations,  $\overrightarrow{h}_{k,j}$ and $\overleftarrow{h}_{k,j}$ are the forward and backward context-dependent representations, and $j=1,...,L$ is the index of the language model's layers. The token representation $x_k$ is computed by a CNN over character embeddings \cite{Kim2016CharacterAwareNL, Jzefowicz2016ExploringTL} and then passed to highway layers \cite{Srivastava2015HighwayN} and a linear projection. The final ELMo representations for the downstream model are derived by collapsing all layers in $R_k$ into a single vector. In this experiment, we select the topmost layer in $R_k$.

\subsection{Transfer Learning}

\begin{figure*} 
  \includegraphics[width=\textwidth]{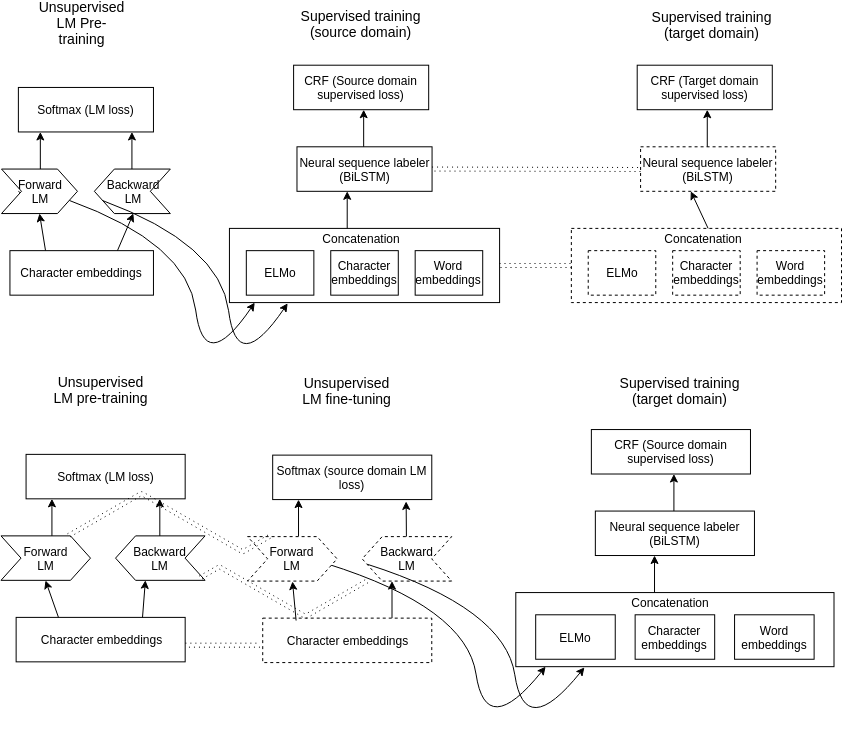}
  \caption{Double dash line represents weight transfer. \textbf{Top}: Supervised fine-tuning by fine-tuning the neural sequence labeler on the target domain. The CRF layer is replaced. \textbf{Bottom}: Unsupervised fine-tuning by fine-tuning the pretrained LM on the source domain.} \label{fig:USS-USS}
\end{figure*}

As mentioned briefly in Section \ref{sec:intro}, there are two variants of transfer learning used in this paper. Details of both approaches are shown in Figure \ref{fig:USS-USS}. 

\paragraph{Supervised fine-tuning}
With ELMo representations, this approach first trains a model using labeled data from source domain $\mathbb{S}$; next, it initializes the target model with the learned parameters; finally, it fine-tunes the target model using labeled data from target domain $\mathbb{T}$. All weights of the model except the pretrained LM are updated.

\paragraph{Unsupervised fine-tuning} 
Unsupervised fine-tuning is inspired by ULMFiT \cite{Ruder2018UniversalLM}. Rather than training on the source domain and fine-tuning on the target domain, this approach fine-tunes the pretrained LM using unlabeled data from source domain $\mathbb{S}$; then, it trains the target model using labeled data from target domain $\mathbb{T}$. All weights of the model except the pretrained LM are updated.

\subsection{Dataset}
In this paper, we use Indonesian language. There are three datasets: unlabeled $\mathbb{U}$, source $\mathbb{S}$, and target $\mathbb{T}$. We evaluate our approach in the settings that $\mathbb{U}$ is of generic domain (newswire or formal) and the $\mathbb{S}$ and $\mathbb{T}$ are of specific domain (conversational). 

For the unlabeled data, we use Kompas-Tempo (newswire) \cite{Tala2003ASO} and Indonesian Wikipedia. Unless stated otherwise, we refer to the former and the latter as \texttt{KT} and \texttt{Idwiki}, respectively. 

For source and target domain data, we use our own manually labeled dataset, the same dataset introduced by
\cite{Kurniawan2018EmpiricalEO}
, namely \texttt{SMALL-TALK} as the target domain and \texttt{TASK-ORIENTED} as the source domain. The former is a 16K conversational messages from users having small talk with a chatbot, whereas the latter contains 12K task-oriented messages such as movie tickets booking, and food delivery. For the rest of the paper, unless stated otherwise, we refer to the former as \texttt{ST} and the latter as \texttt{TO}. 

For the unsupervised fine-tuning approach, since we do not need labeled data for the source domain $\mathbb{S}$, we can easily add more unlabeled dataset to our source domain $\mathbb{S}$. Conveniently, we have a large unlabeled Indonesian conversational texts at our disposal, which is a superset of the \texttt{TO} labeled dataset. We refer to this bigger unlabeled conversational dataset as \texttt{TOL}. Using this dataset, we can perform unsupervised fine-tuning with a larger data size compared to that of supervised fine-tuning. 

Due to proprietary and privacy reasons, unfortunately, we cannot publish our Indonesian conversational texts. We present our dataset statistics in Table \ref{tab:indo-stats} and the label details in Table \ref{tab:label-detail-st} and \ref{tab:label-detail-to}.

  
  

\begin{table}
\parbox{.45\linewidth}{
\centering
  \small
  \caption{Number of sentences (\textit{N}), number of distinct words (\textit{DW}), number of tokens (\textit{T}), and average sentence length (\textit{AVG}) in \texttt{TO} , \texttt{ST}, and \texttt{TOL} dataset.} \label{tab:indo-stats}
  \begin{tabular}{ccccc}
    \toprule
    {} & \textit{N} & \textit{DW} & \textit{T} & \textit{AVG} \\ 
    \midrule
    \texttt{TO} train & 10142 & 16930 & 129841 & 12.7 \\
    \texttt{TO} dev & 1250 & 4291 & 16021 & 12.8  \\
    \texttt{TO} test & 1205 & 3868 & 14610 & 12.12  \\
    \texttt{ST} train & 11577 & 9034 & 434408 & 3.74 \\
    \texttt{ST} dev & 3289 & 3749 & 12583 & 3.82 \\
    \texttt{ST} test & 1641 & 2265 & 6300 & 3.83 \\
    \texttt{TOL} & 916838 & 144028 & 6652122 & 7.25 \\
    \bottomrule
  \end{tabular}
}
\hfill
\parbox{.45\linewidth}{
\centering
  \small
  \caption{Number of entities contained in the \texttt{ST} dataset. There are 6 labels in total.} \label{tab:label-detail-st}
  \begin{tabular}{cccc}
    \toprule
    \textbf{Entity} & \textbf{Train} & \textbf{Test} & \textbf{Dev} \\ 
    \midrule
    DATETIME & 49 & 20 & 21 \\
    EMAIL & 20 & 7 & 8 \\
    GENDER & 241 & 64 & 85 \\
    LOCATION & 2672 & 813 & 867 \\
    PERSON & 2455 & 749 & 754 \\
    PHONE & 44 & 18 & 21 \\ 
    \bottomrule
  \end{tabular}
}
\end{table}

\begin{table}[!htbp]
  \centering
  \caption{Number of entities contained in the \texttt{TO} dataset. There are 13 labels in total.} \label{tab:label-detail-to}
  \resizebox{\textwidth}{!}{
  \begin{tabular}{cccccccccccccc}
    \toprule
    \textbf{Entity} & AREA & CURRENCY & DATETIME & DURATION & EMAIL & LENGTH & LOCATION & NUMBER & PERSON & PHONE & TEMPERATURE & VOLUME & WEIGHT \\ 
    \midrule
    \textbf{Train} & 78 & 1213 & 2982 & 339 & 226 & 169 & 6322 & 3966 & 4031 & 466 & 70 & 58 & 134 \\
    \textbf{Dev} & 6 & 139 & 350 & 38 & 23 & 20 & 791 & 471 & 480 & 53 & 8 & 8 & 11 \\
    \textbf{Test} & 6 & 140 & 347 & 26 & 21 & 14 & 709 & 404 & 477 & 49 & 4 & 4 & 11 \\
    \bottomrule
  \end{tabular}}
\end{table}


\section{Experiments and results} \label{sec:res}
\subsection{Experiment setup} 
We use Rei et al.'s \cite{Rei2017SemisupervisedML} 
implementation\footnote{\url{https://github.com/marekrei/sequence-labeler}} for the multitask model and AllenNLP\footnote{\url{https://github.com/allenai/allennlp}} for the rest of our models. For every combination of training dataset and model, we tune the dropout rate \cite{Srivastava2014DropoutAS} by grid search on [0.25, 0.35, 0.5, 0.65, 0.75] for both the ELMo and the neural sequence labeler using the same random seed for all configurations. We do not tune other hyperparameters due to computational resource constraints. 

For all models, we use identical data pre-processing: words are lowercased, BIO scheme is used, and start end tokens are excluded from the vocabulary. For the multitask model, we use identical settings to that of \cite{Rei2017SemisupervisedML}. For the rest of the models, we use different settings: word and character embedding sizes are set to 50 and 16, respectively. Character embeddings are formed by a CNN with 128 filters followed by highway layers and a ReLU activation layer. Both word and character embeddings are initialized randomly. ELMo embedding size is set to 1024. The LSTMs are set to have 200 hidden units and 2 layers. We apply L2 regularizer of 0.1 to all layers and early stopping is used with a patience of 25. We use Adam \cite{Kingma2014AdamAM} optimization with learning rate of 0.001 and we clip the gradient at 5.0. We run our experiments with batch size of 32 and epochs of 150. We evaluated all our experiments with CoNLL evaluation: micro-averaged F1 score based on exact span matching.

\subsection{Impact of pretrained LM embeddings} 
To assess the impact of the pretrained LM embeddings, first we compare the performance of the baseline model to the baseline models with pretrained LM embeddings on NER task evaluated on \texttt{TO} dataset. We use CNN-BiLSTM-CRF by \cite{Ma2016EndtoendSL} as our baseline model and ELMo as our pretrained LM embeddings. In addition to ELMo, we also use Flair \cite{akbik2018coling} as it reaches state-of-the-art performance on English NER task\footnote{as of 03/01/2019}. Flair also differs from ELMo because it is trained with character-level softmax. We use the default hyperparameters provided by \cite{akbik2018coling} to train Flair LM\footnote{\url{https://github.com/zalandoresearch/flair}}.

\begin{table}  
  \centering
  \caption{Accuracy of the baseline model and the baselines with the pretrained embeddings (ELMo and Flair trained on \texttt{Idwiki} and \texttt{KT} data) trained and evaluated on \texttt{TO} data.} \label{tab:impact-LM}
  \begin{tabular}{ccc}
    \toprule
    \textbf{Models} & \textbf{F1 Dev} & \textbf{F1 Test} \\ 
    \midrule
    CNN-BiLSTM-CRF & 85.94 & 85.85  \\ 
    +ELMo \texttt{Idwiki} & 87.67 & 87.72  \\ 
    +ELMo \texttt{KT} & 86.88 & 87.11  \\ 
    +Flair \texttt{Idwiki} & 87.12 & 88.41  \\ 
    +Flair \texttt{KT} & 71.16 & 71.73 \\ 
    \bottomrule
  \end{tabular}
\end{table}

From Table \ref{tab:impact-LM} we can see that adding LM embeddings improves the overall performance. ELMo yields small absolute improvement of dev F1 score when trained on both \texttt{Idwiki} and \texttt{KT} than the baseline. Flair, although obtains roughly the same dev F1 score when trained on \texttt{Idwiki}, does not perform well when trained on \texttt{KT}. This result might be attributed to the fact that \texttt{KT} contains far fewer sentences compared to \texttt{IdWiki} (around 9 times fewer). We link this result to an observation made by \cite{Yu2018OnTS} which stated that character language models are unstable when the training data is not big enough. Therefore, we do not proceed with Flair for the next experiment with our two approaches.

\subsection{Main experiment} 
We experiment with three model groups. The first group is the baseline models, consisting of a single-task learning model using CNN-BiLSTM-CRF \cite{Ma2016EndtoendSL} and a multitask approach \cite{Rei2017SemisupervisedML}, which uses an LSTM-BiLSTM-CRF with an additional LM loss. The second and third groups involve our first and second approaches, which are the unsupervised language model fine-tuning and the supervised neural sequence labeler fine-tuning, respectively. 

We train the models in each group on different percentages of the target domain training data $\mathbb{T}$. We do this to assess the impact of our supervised and unsupervised fine-tuning approach when presented with varying amount of training data. Table \ref{tab:st-varying} shows the number of entities of the \texttt{ST} data with different percentages of training sentences used. Table \ref{tab:fin-result} shows the result of our experiments. All numbers shown in the table come from models with the best hyperparameter on the target domain validation set.

\begin{table*}  
  \centering
  \small
  \caption{Number of entities contained in the \texttt{ST} training dataset. The training data percentage is followed by the total number of training sentences.} \label{tab:st-varying}
  \begin{tabular}{ccccccc}
    \toprule
    \textbf{Entity} & \textbf{1\%-100} &
    \textbf{5\%-502} & \textbf{10\%-1004} & \textbf{25\%-2511} & \textbf{50\%-5022} & \textbf{75\%-7533} \\ 
    \midrule
    DATETIME & 2 & 4 & 8 & 18 & 27 & 39 \\
    EMAIL & 2 & 5 & 3 & 7 & 11 & 17 \\
    GENDER & 4 & 18 & 26 & 65 & 123 & 174 \\
    LOCATION & 30 & 137 & 265 & 640 & 1345 & 2019 \\
    PERSON & 20 & 122 & 251 & 606 & 1216 & 1829 \\
    PHONE & 1 & 2 & 5 & 8 & 22 & 31 \\
    \bottomrule
  \end{tabular}
\end{table*}

\begin{table*}[!htbp]
    \centering
    \caption{Experiment results on the \texttt{ST} test data. Models are trained on \texttt{ST} training data with varying number of training instances. Bold and underline indicate the highest and the second highest test F1 score, respectively.} \label{tab:fin-result}
    \resizebox{\textwidth}{!}{
    \begin{tabular}{ccccccccccc}
    \toprule
    \multirow{2}{*}{Group} & \multirow{2}{*}{Model name} & Unlabeled data for & Source domain & \multicolumn{7}{c}{Target domain test F1 score} \\
    {} & {} & pretrained LM & data & 1\%  & 5\% & 10\% & 25\% & 50\% & 75\% & 100\% \\
    \midrule
    \multirow{2}{*}{Baselines} & CNN-BiLSTM-CRF & - & - & 0.59 & 43.84 & 50.86 & 62.19 & 72.64 & 72.79 & 75.81 \\
    {} & Rei (2017) & - & - & 41.44 & 59.43 & 68.03 & 74.30 & 80.47 & \underline{83.23} & 85.12 \\ \hline
    \multirow{2}{*}{Supervised fine-tuning} & LM[Idwiki]\_Sup[TO-ST] & \texttt{Idwiki} & \texttt{TO} & \textbf{73.17} & \textbf{77.25} & 77.21 & \underline{80.31} & 82.27 & \textbf{83.54} & 84.62 \\
    {} & LM[KT]\_Sup[TO-ST] & \texttt{KT} & \texttt{TO} & \underline{71.84} & 75.16 & 76.42 & 79.11 & \underline{82.88} & 83.18 & 84.49 \\ \hline
    \multirow{2}{*}{Unsupervised fine-tuning} & LM[Idwiki-TOL]\_Sup[ST] & \texttt{Idwiki} & \texttt{TOL} & 67.10 & 74.05 & \underline{77.49} & 79.76 & \textbf{83.82} & 83.17 & \textbf{85.78} \\
    {} & LM[KT-TOL]\_Sup[ST] & \texttt{KT} & \texttt{TOL} & 64.48 & \underline{75.55} & \textbf{77.77} & \textbf{80.99} & 83.09 & 82.74 & \underline{85.76} \\ 
    \bottomrule
    \end{tabular}}
\end{table*}

\subsubsection{Model name conventions}
Here we explain the patterns for naming our models in Table \ref{tab:fin-result} and the rest of our paper: LM indicates an unsupervised step, whereas Sup is for the supervised step. A dash (-) shows a fine-tuning step (supervised or unsupervised), an underscore (\_) represents the move from LM training to BiLSTM training, and a square bracket ([]) represents which dataset used in the unsupervised (LM) or the supervised (Sup) step. For example, one can interpret a model named LM[Idwiki]\_Sup[TO-ST] as a model which: (1) uses pretrained LM trained on \texttt{Idwiki} dataset, (2) trains the BiLSTM on the source domain dataset (\texttt{TO}), and (3) fine-tunes the BiLSTM on the target domain dataset (\texttt{ST}). 

\subsubsection{Results discussion}

\paragraph{Baseline and multitask} Unsurprisingly, CNN-BiLSTM-CRF fails when the training data is as small as 100 sentences, reaching only 0.59\% test F1 score. As the training data gradually increases, CNN-BiLSTM-CRF performs fairly well, reaching 75.81\% test F1 score on the whole data. The multi-task model by \cite{Rei2017SemisupervisedML} reaches 45.10\% test F1 score when trained on 100 sentences training data. The absolute improvement from the baseline is huge considering that it only benefits from the LM loss and no additional signals from external labeled/unlabeled data are incorporated. It is also competitive compared to other models when trained on the whole data. 

\paragraph{Supervised fine-tuning} There is a significant increase of test F1 score using supervised fine-tuning compared to the baseline on every level of training data size.
The gain from supervised fine-tuning is largest when the data is small. Both LM[Idwiki]\_Sup[TO-ST] and LM[KT]\_Sup[TO-ST] achieve the highest and second highest test F1 score when trained on 1\% training data. 
However, when trained on 100\% training data, supervised fine-tuning does not perform better than the multitask model. The gain of supervised fine-tuning seems to be diminishing as the training data grows larger. 
This hints that using supervised fine-tuning might not be necessary if labeled data is already present in adequate amount thus one can opt for simpler models such as the multitask model.

\paragraph{Unsupervised fine-tuning} Overall, using unsupervised fine-tuning yields competitive result with the supervised fine-tuning. A notable difference is  when the model trained with 1\% and 100\% training data. LM[Idwiki-TOL]\_Sup[ST] and LM[KT-TOL]\_Sup[ST] obtains the highest and the second highest F1 test score on 100\% training data, respectively. On 1\% training data, they fall behind the supervised fine-tuning by about 5 points of test F1 score. The huge unlabeled source domain data for the LM fine-tuning does not seem to be helping much on small training data. This hints that labeled data in small-moderate size is still more effective compared to unlabeled data which comes in massive size. Also, there does not seem to be a significant difference between the unsupervised fine-tuning and the multitask approach on 75\% and 100\% training data. Multitask approach seems to perform competitively when the training data size is vast enough. Again, this tells us that we might not need to perform unsupervised fine-tuning if labeled data is already present in adequate amount. 

\subsubsection{Ablation}
An ablation study is conducted for both the supervised and unsupervised fine-tuning. We perform three ablations resulting in four models: two for the supervised fine-tuning group and one for the  unsupervised fine-tuning. We carry out the ablations on the development set. Table \ref{tab:ablation} shows the result of the ablated models compared with the models of our two approaches.

\begin{table*}[!htbp]
    \centering
    \caption{Experiment results on the \texttt{ST} dev data. The original models and the ablation models are trained on \texttt{ST} training data with varying number of training instances. Bold and underline indicates the highest and the second highest test F1 score, respectively.} \label{tab:ablation}
    \resizebox{\textwidth}{!}{
    \begin{tabular}{ccccccccccc}
    \toprule
    \multirow{2}{*}{Group} & \multirow{2}{*}{Model name} & Unlabeled data for & Source domain & \multicolumn{7}{c}{Target domain dev F1 score} \\
    {} & {} & pretrained LM & data & 1\%  & 5\% & 10\% & 25\% & 50\% & 75\% & 100\% \\
    \midrule
    \multirow{2}{*}{Supervised fine-tuning} & LM[Idwiki]\_Sup[TO-ST] & \texttt{Idwiki} & \texttt{TO} & \textbf{71.12} & 75.02 & 77.28 & 80.36 & 81.85 & 81.82 & 82.96 \\
    {} & LM[KT]\_Sup[TO-ST] & \texttt{KT} & \texttt{TO} & \underline{70.42} & 73.79 & 75.12 & 79.95 & 81.39 & 82.04 & 82.33 \\ \hline
    \multirow{2}{*}{Unsupervised fine-tuning} & LM[Idwiki-TOL]\_Sup[ST] & \texttt{Idwiki} & \texttt{TOL} & 65.43 & 74.64 & 77.71 & 81.14 & 82.68 & \underline{83.71} & \underline{84.37} \\
    {} & LM[KT-TOL]\_Sup[ST] & \texttt{KT} & \texttt{TOL} & 63.90 & \textbf{75.75} & \underline{77.84} & \textbf{82.38} & \underline{83.75} & 83.26 & \textbf{85.11} \\ \hline
    \multirow{4}{*}{Ablation} & Sup[TO-ST] & - & \texttt{TO} & 66.08 & 69.94 & 71.91 & 76.68 & 81.00 & 80.78 & 82.04 \\
    {} & LM[Idwiki]\_Sup[ST] & \texttt{Idwiki} & - & 55.99 & 69.15 & 71.94 & 74.92 & 79.60 & 80.31 & 80.71 \\
    {} & LM[KT]\_Sup[ST] & \texttt{KT} & - & 53.79 & 66.70 & 71.39 & 74.61 & 78.68 & 78.91 & 81.02 \\ 
    {} & LM[TOL]\_Sup[ST] & \texttt{TOL} & - & 61.59 & \underline{75.05} & \textbf{78.60} & \underline{82.00} & \textbf{83.91} & \textbf{85.04} & 84.18 \\
    \bottomrule 
    \end{tabular}}
\end{table*}

The first ablation is removing the LM pretraining step but keeping the supervised fine-tuning using the source domain (Sup[TO-ST]). Notice that there is a sizable drop of test F1 score on every training data size compared to the supervised fine-tuning models, especially on 1\% training data. The second ablation is omitting the intermediate supervised fine-tuning step while keeping the pretrained LM intact (LM[Idwiki]\_Sup[ST] and LM[KT]\_Sup[ST]). Without the supervised fine-tuning, the models' performance is markedly curtailed. 
Another interesting pattern is that removing the LM pretraining step does less harm than removing the supervised fine-tuning step, especially on small training data. On 1\% training data, Sup[TO-ST] outperforms LM[Idwiki]\_Sup[ST] and LM[KT]\_Sup[ST] by significant margins.
It seems that the supervised fine-tuning part is where the model benefits the most. Nonetheless, even without the supervised fine-tuning, the pretrained LM alone is already a huge reinforcement for the models.

The last ablation is conducted by excluding the unsupervised fine-tuning step, which is the model named LM[TOL]\_Sup[ST]. We do the LM pretraining directly with the source domain data (\texttt{TOL}) without using the generic domain data.
LM[TOL]\_Sup[ST] obtains highest F1 test score on 10\%, 50\%, and 75\% training data and second highest F1 test score on 5\% and 25\% training data. This is quite surprising considering that LM[TOL]\_Sup[ST] is trained with the same fashion as LM[Idwiki]\_Sup[ST] and LM[KT]\_Sup[ST]: no fine-tuning and only BiLSTM with LM pretraining. The only difference is in the dataset used for the LM pretraining. We hypothesize that the gain stems from the fact that \texttt{TOL} is of conversational domain, whereas \texttt{Idwiki} and \texttt{KT} is of generic domain. Overall, LM[TOL]\_Sup[ST] also performs competitively compared to both the supervised and unsupervised fine-tuning models on every level of training data size. This highlights that one might not need to perform fine-tuning from a generic domain if a large in-domain unlabeled data is already at hand.
\section{Analysis} \label{sec:anal}
From the observations of the previous section, we conclude that there are three main highlights regarding our fine-tuning approaches: (1) supervised transfer performs best on 1\% training data, (2) unsupervised transfer obtains the best F1 score on 100\% training data, and (3) pretraining the LM directly on the source domain (\texttt{TOL}) without fine-tuning works really well, attaining comparable result with both fine-tuning approaches. Here, we seek to establish more understanding of why (1), (2), and (3) happen.

Recent research have discovered that pretrained LM induce useful knowledge such as syntactic information for downstream tasks \cite{Blevins2018DeepRE, Linzen2016AssessingTA, Gulordava2018ColorlessGR}. Motivated by these findings, we formulate a hypothesis that the supervised and unsupervised fine-tuning models also learn advantageous knowledge for predicting named entities, namely part-of-speech (POS) information. It is also known that POS has been used as features for NER and they help improve the overall performance \cite{Curran2003LanguageIN}. To test this hypothesis, we train a diagnostic classifier \cite{Veldhoen2016DiagnosticCR, Hupkes2018VisualisationA} using the models' hidden state representations as features on an Indonesian part-of-speech conversational dataset. In this experiment, the diagnostic classifier is a simple classifier (single layer neural network with a softmax output) to predict the POS tags. During training, the weights of the models are all frozen except the diagnostic classifier's weights. To train the diagnostic classifier, we use our own manually labeled POS dataset because to the best of our knowledge, there is no labeled Indonesian part-of-speech conversational dataset that is publicly available. We also cannot publish our POS dataset due to proprietary and privacy reasons. We provide the dataset statistics in Table \ref{tab:pos-stats} and the label details in Table \ref{tab:pos-labels}.

\paragraph{POS tag schema}
Here we explain the schema used in our POS tag dataset. The dataset is annotated by an Indonesian linguist. All 21 tags shown in Table \ref{tab:pos-labels} are the same tags as the English Penn Treebank POS tags \cite{Marcus:1994:PTA:1075812.1075835} except for: (1) CDI, NEG, PRL, SC, VBI, and VBT, which were taken from the Indonesian POS Tagset 1 \cite{Pisceldo2009ProbabilisticPO} and (2) NUM, PNP, and X, which were created based on the Indonesian grammar references. NUM stands for numbers (e.g., \textit{tujuh}-seven). PNP is number pronouns, which is used to refer to person or object identified with numbers (e.g., \textit{keduanya}-the two of them). X is for unrecognized words (e.g., \textit{wkwk}-wkwk).

\begin{table}[!htbp]
  \centering
  \caption{Number of sentences (\textit{N}), number of distinct words (\textit{DW}), number of tokens (\textit{T}), and average sentence length (\textit{AVG}) in POS dataset.} \label{tab:pos-stats}
  \begin{tabular}{ccccc}
    \toprule
    {} & \textit{N} & \textit{DW} & \textit{T} & \textit{AVG} \\ 
    \midrule
    POS train & 4108 & 7077 & 34843 & 8.48 \\
    POS dev & 1008 & 2109 & 9115 & 9.04  \\
    POS test & 1283 & 3209 & 11111 & 8.66  \\
    \bottomrule
  \end{tabular}
\end{table}

\begin{table}[!htbp]
  \centering
  \caption{Number of labels contained in our POS dataset. There are 23 labels in total.} \label{tab:pos-labels}
  \resizebox{\textwidth}{!}{
  \begin{tabular}{cccccccccccccccccccccccc}
    \toprule
    \textbf{Entity} & CC & CDI & DT & FW & IN & JJ & MD & NEG & NN & NNP & NUM & PNP & PRL & PRP & RB & RP & SC & SYM & UH & VBI & VBT & WP & X \\ 
    \midrule
    \textbf{Train} & 412 & 73 & 1120 & 1679 & 1493 & 750 & 1126 & 447 & 7236 & 5213 & 1178 & 2 & 27 & 1018 & 883 & 193 & 1126 & 4787 & 1589 & 892 & 2817 & 722 & 60 \\
    \textbf{Dev} & 110 & 17 & 295 & 561 & 439 & 182 & 271 & 110 & 1857 & 1341 & 280 & 1 & 7 & 247 & 241 & 42 & 306 & 1207 & 382 & 235 & 807 & 169 & 8 \\
    \textbf{Test} & 166 & 25 & 350 & 505 & 502 & 238 & 365 & 142 & 2306 & 1652 & 370 & 2 & 12 & 337 & 307 & 61 & 339 & 1481 & 511 & 286 & 924 & 216 & 14 \\
    \bottomrule
  \end{tabular}}
\end{table}

We train the diagnostic classifier using the hidden representations of: (1) the pretrained LM (for both the supervised and unsupervised fine-tuning approaches), (2) the neural sequence labeler (BiLSTM) trained on the source domain (for supervised fine-tuning), and (3) the fine-tuned language model on the source domain (for unsupervised fine-tuning). We report the accuracy on the development set. The test is carried out to check which step encodes the part-of-speech information better for both the supervised and unsupervised fine-tuning models. Note that here there is no target domain involved since we only want to know the quality of the POS information learned from the fine-tuning step.

\begin{table*}[!htbp]
    \centering
    \caption{Diagnostic classifiers trained on part-of-speech (POS) task using the different models' hidden representations as input. We present the accuracy on the development set.} \label{tab:pos-result}
    \resizebox{\textwidth}{!}{
    \begin{tabular}{ccccc}
    \toprule
    Hidden representations input & Model name & Pretrained LM data & Source domain & Dev acc \\
    \midrule
    - & Most frequent tag & - & - & 84.51 \\
    \multirow{3}{*}{Pretrained LM} & LM[Idwiki] & \texttt{Idwiki} & - & 86.14\\
    {} & LM[KT] & \texttt{KT} & - & 86.68 \\
    {} & LM[TOL] & \texttt{TOL} & - & 92.05  \\
    \hline
    \multirow{2}{*}{Fine-tuned LM} & LM[Idwiki-TOL] & \texttt{Idwiki} & \texttt{TOL} & 92.52  \\
    {} & LM[KT-TOL] & \texttt{KT} & \texttt{TOL} & 93.08 \\
    \hline
    \multirow{3}{*}{Neural sequence labeler (BiLSTM)} & LM[Idwiki]\_Sup[TO] & \texttt{Idwiki} & \texttt{TO} & 63.64 \\
    {} & LM[KT]\_Sup[TO] & \texttt{KT} & \texttt{TO} & 63.83 \\
    {} & Sup[TO] & - & \texttt{TO} & 58.62 \\
    \bottomrule
    \end{tabular}}
\end{table*}

Table \ref{tab:pos-result} shows the result of the diagnostic classifier trained on our POS dataset. In addition to the diagnostic classifier, we provide a simple baseline model that outputs the most frequent tag in the training set for a given word. For OOV word, the simple baseline model outputs the most frequent label in the training data.
Given inputs from the pretrained LM, the diagnostic classifier performs considerably well on the development set. LM[TOL], LM[Idwiki], and LM[KT] obtains higher accuracies than the simple baseline model. LM[TOL] reaches higher accuracy than LM[Idwiki] and LM[KT]. This result aligns with our previous result that \texttt{TOL} is of conversational domain hence it is more useful for downstream tasks in conversational domain. The diagnostic classifier reaches slightly better accuracy when trained using inputs from the fine-tuned LM, yielding an absolute improvement of 1 point of dev accuracy. Since the differences between the accuracies of the diagnostic classifier are minuscule, this may explain why LM pretraining without fine-tuning (LM[TOL]\_Sup[ST]) is competitive with its fine-tuning counterpart (LM[Idwiki-TOL]\_Sup[ST] and LM[Idwiki-KT]\_Sup[ST]). Unsupervised fine-tuning might not be necessary if one already has access to a huge unlabeled in-domain data. With this result, we can also conclude that the LM pretraining (ELMo) induces useful syntactic knowledge, which in this case is part-of-speech information.

Surprisingly, the diagnostic classifier performances badly deteriorate on the development set given inputs from the BiLSTM. All models from this group obtain accuracies below the simple baseline model. This contradicts our previous result where supervised fine-tuning obtains adequate result on small training data. It seems that the BiLSTM does not encode part-of-speech information as good as the pretrained LM, even though it receives additional input from the pretrained LM (LM[Idwiki]\_Sup[TO] and LM[KT]\_Sup[TO]). We think that the supervised fine-tuning models may learn something other than the part-of-speech information which helps them perform well on the small training data. A plausible explanation would be that BiLSTM is learning NER-specific information during the supervised training on source domain, replacing the part-of-speech information from the pretrained LM.
\section{Conclusion} \label{sec:con}
In this paper, we investigate the impact of language model pretraining on named-entity recognition task in Indonesian conversational texts. We use two variants of three step training procedure: supervised fine-tuning (fine-tuning the BiLSTM) and unsupervised fine-tuning (fine-tuning the pretrained LM). Using both approaches, the neural models obtain significant increase from the CNN-BiLSTM-CRF and the multitask baseline on small training data, yielding an absolute improvement of 32 points of test F1 score. 
However, one might not need to fine-tune if: (1) a large unlabeled in-domain data is already at hand then one can train language model directly without any fine-tuning, and (2) an adequate amount of labeled in-domain data (in our case it's $> 5000$ sentences) is present then one can opt for simpler models such as the multitask model. 
Furthermore, we also find that the pretrained LM encodes part-of-speech information, which is a strong predictor for named entity recognition. The neural sequence labeler, on the other hand, seems to encode another information other than part-of-speech to help it perform well on NER task on small training data.

%
%

\bibliographystyle{splncs04}
\bibliography{main}

\begin{thebibliography}{10}
\providecommand{\url}[1]{\texttt{#1}}
\providecommand{\urlprefix}{URL }
\providecommand{\doi}[1]{https://doi.org/#1}

\bibitem{Aguilar2017AMA}
Aguilar, G., Maharjan, S., L{\'o}pez-Monroy, A.P., Solorio, T.: A multitask
  approach for named entity recognition in social media data (2017)

\bibitem{akbik2018coling}
Akbik, A., Blythe, D., Vollgraf, R.: Contextual string embeddings for sequence
  labeling. In: {COLING} 2018, 27th International Conference on Computational
  Linguistics. pp. 1638--1649 (2018)

\bibitem{Aryoyudanta2016SemisupervisedLA}
Aryoyudanta, B., Adji, T.B., Hidayah, I.: Semi-supervised learning approach for
  indonesian named entity recognition (ner) using co-training algorithm. 2016
  International Seminar on Intelligent Technology and Its Applications (ISITIA)
  pp. 7--12 (2016)

\bibitem{Blevins2018DeepRE}
Blevins, T., Levy, O., Zettlemoyer, L.S.: Deep rnns encode soft hierarchical
  syntax. In: ACL (2018)

\bibitem{Budi2005NIL}
Budi, I., Bressan, S., Wahyudi, G., Hasibuan, Z.A., Nazief, B.A.A.: Named
  entity recognition for the indonesian language: Combining contextual,
  morphological and part-of-speech features into a knowledge engineering
  approach. In: Hoffmann, A., Motoda, H., Scheffer, T. (eds.) Discovery
  Science. pp. 57--69. Springer Berlin Heidelberg, Berlin, Heidelberg (2005)

\bibitem{Chiticariu2010DomainAO}
Chiticariu, L., Krishnamurthy, R., Li, Y., Reiss, F., Vaithyanathan, S.: Domain
  adaptation of rule-based annotators for named-entity recognition tasks. In:
  EMNLP (2010)

\bibitem{Curran2003LanguageIN}
Curran, J.R., Clark, S.: Language independent ner using a maximum entropy
  tagger. In: CoNLL (2003)

\bibitem{Daum2007FrustratinglyED}
Daum{\'e}, H.: Frustratingly easy domain adaptation. CoRR
  \textbf{abs/0907.1815} (2007)

\bibitem{Gulordava2018ColorlessGR}
Gulordava, K., Bojanowski, P., Grave, E., Linzen, T., Baroni, M.: Colorless
  green recurrent networks dream hierarchically. In: NAACL-HLT (2018)

\bibitem{Hupkes2018VisualisationA}
Hupkes, D., Veldhoen, S., Zuidema, W.H.: Visualisation and 'diagnostic
  classifiers' reveal how recurrent and recursive neural networks process
  hierarchical structure. In: IJCAI (2018)

\bibitem{Jzefowicz2016ExploringTL}
J{\'o}zefowicz, R., Vinyals, O., Schuster, M., Shazeer, N., Wu, Y.: Exploring
  the limits of language modeling. CoRR  \textbf{abs/1602.02410} (2016)

\bibitem{Kim2016CharacterAwareNL}
Kim, Y., Jernite, Y., Sontag, D.A., Rush, A.M.: Character-aware neural language
  models. In: AAAI (2016)

\bibitem{Kingma2014AdamAM}
Kingma, D.P., Ba, J.: Adam: A method for stochastic optimization. CoRR
  \textbf{abs/1412.6980} (2014)

\bibitem{Kurniawan2018EmpiricalEO}
Kurniawan, K., Louvan, S.: Empirical evaluation of character-based model on
  neural named-entity recognition in indonesian conversational texts. CoRR
  \textbf{abs/1805.12291} (2018)

\bibitem{Lake2017StillNS}
Lake, B.M., Baroni, M.: Still not systematic after all these years: On the
  compositional skills of sequence-to-sequence recurrent networks. CoRR
  \textbf{abs/1711.00350} (2017)

\bibitem{Lee2018TransferLF}
Lee, J.Y., Dernoncourt, F., Szolovits, P.: Transfer learning for named-entity
  recognition with neural networks. CoRR  \textbf{abs/1705.06273} (2018)

\bibitem{Leonandya2015ASA}
Leonandya, R., Distiawan, B., Praptono, N.H.: A semi-supervised algorithm for
  indonesian named entity recognition. 2015 3rd International Symposium on
  Computational and Business Intelligence (ISCBI) pp. 45--50 (2015)

\bibitem{Linzen2016AssessingTA}
Linzen, T., Dupoux, E., Goldberg, Y.: Assessing the ability of lstms to learn
  syntax-sensitive dependencies. TACL  \textbf{4},  521--535 (2016)

\bibitem{Luthfi2014BuildingAI}
Luthfi, A., Trisedya, B.D., Manurung, R.: Building an indonesian named entity
  recognizer using wikipedia and dbpedia. 2014 International Conference on
  Asian Language Processing (IALP) pp. 19--22 (2014)

\bibitem{Ma2016EndtoendSL}
Ma, X., Hovy, E.H.: End-to-end sequence labeling via bi-directional
  lstm-cnns-crf. CoRR  \textbf{abs/1603.01354} (2016)

\bibitem{Marcus:1994:PTA:1075812.1075835}
Marcus, M., Kim, G., Marcinkiewicz, M.A., MacIntyre, R., Bies, A., Ferguson,
  M., Katz, K., Schasberger, B.: The penn treebank: Annotating predicate
  argument structure. In: Proceedings of the Workshop on Human Language
  Technology. pp. 114--119. HLT '94, Association for Computational Linguistics,
  Stroudsburg, PA, USA (1994). \doi{10.3115/1075812.1075835},
  \url{https://doi.org/10.3115/1075812.1075835}

\bibitem{Peters2018DeepCW}
Peters, M.E., Neumann, M., Iyyer, M., Gardner, M., Clark, C., Lee, K.,
  Zettlemoyer, L.S.: Deep contextualized word representations. In: NAACL-HLT
  (2018)

\bibitem{Pisceldo2009ProbabilisticPO}
Pisceldo, F.: Probabilistic part of speech tagging for bahasa indonesia (2009)

\bibitem{Radford2018ImprovingLU}
Radford, A.: Improving language understanding by generative pre-training (2018)

\bibitem{Rei2017SemisupervisedML}
Rei, M.: Semi-supervised multitask learning for sequence labeling. In: ACL
  (2017)

\bibitem{Ruder2018UniversalLM}
Ruder, S., Howard, J.: Universal language model fine-tuning for text
  classification. In: ACL (2018)

\bibitem{Srivastava2014DropoutAS}
Srivastava, N., Hinton, G.E., Krizhevsky, A., Sutskever, I., Salakhutdinov, R.:
  Dropout: a simple way to prevent neural networks from overfitting. Journal of
  Machine Learning Research  \textbf{15},  1929--1958 (2014)

\bibitem{Srivastava2015HighwayN}
Srivastava, R.K., Greff, K., Schmidhuber, J.: Highway networks. CoRR
  \textbf{abs/1505.00387} (2015)

\bibitem{Suwarningsih2014ImNERIM}
Suwarningsih, W., Supriana, I., Purwarianti, A.: Imner indonesian medical named
  entity recognition. 2014 2nd International Conference on Technology,
  Informatics, Management, Engineering and Environment pp. 184--188 (2014)

\bibitem{Tala2003ASO}
Tala, F.Z.: A study of stemming effects on information retrieval in bahasa
  indonesia (2003)

\bibitem{Taufik2016NamedER}
Taufik, N., Wicaksono, A.F., Adriani, M.: Named entity recognition on
  indonesian microblog messages. 2016 International Conference on Asian
  Language Processing (IALP) pp. 358--361 (2016)

\bibitem{Veldhoen2016DiagnosticCR}
Veldhoen, S., Hupkes, D., Zuidema, W.H.: Diagnostic classifiers revealing how
  neural networks process hierarchical structure. In: CoCo@NIPS (2016)

\bibitem{Wu2009DomainAB}
Wu, D., Lee, W.S., Ye, N., Chieu, H.L.: Domain adaptive bootstrapping for named
  entity recognition. In: EMNLP (2009)

\bibitem{Yu2018OnTS}
Yu, X., Mayhew, S.D., Sammons, M., Roth, D.: On the strength of character
  language models for multilingual named entity recognition. In: EMNLP (2018)

\end{thebibliography}

\end{document}